\documentclass[11pt,a4paper]{article}
\usepackage[T1]{fontenc}
\usepackage[utf8]{inputenc}
\usepackage{lmodern}
\usepackage[margin=2.6cm]{geometry}
\usepackage{amsmath,amssymb,amsthm}
\usepackage{graphicx}
\usepackage{booktabs}
\usepackage{tikz}
\usetikzlibrary{arrows.meta,positioning,fit}
\usepackage[hidelinks]{hyperref}
\usepackage{microtype}

\newtheorem{proposition}{Proposition}
\newtheorem{lemma}{Lemma}
\theoremstyle{remark}

\newcommand{\dd}{\,\mathrm{d}}

\title{Neural networks whose nonlinearity lives in the memory:\\
linear general fractional systems with state-dependent order}
\author{Stefano Fazzino\\[2pt]
\small Independent researcher, Italy \quad ORCID: 0000-0002-3015-4938}
\date{}

\begin{document}
\maketitle

\begin{abstract}
In recurrent neural networks the nonlinearity is usually placed in the vector field: an activation
function, a learned right-hand side, or gates that rescale the state, the input and the time step. This
paper studies networks in which the vector field is linear and the nonlinearity is confined to a single
scalar per equation: the order of differentiation, which depends on the input and on the state. Each unit
is a set of linear equations written with the general fractional derivatives of Kochubei and Luchko, the
units are coupled linearly through a fixed matrix, and the network computes only by choosing, at every
instant, the law with which each equation remembers its input. We give the model a precise meaning
through the diffusive representation, in which a fractional operator becomes a continuum of relaxing
modes, and we require every memory kernel to be a Sonine kernel, so that each equation with frozen order
is a genuine general fractional equation. Three structural facts follow. The equilibria do not depend on
the orders, so all nonlinear computation happens in transients. The linearisation at an equilibrium is
the linear system with the orders frozen at that point. And, for a pure fractional kernel, the order acts
on the response mainly as a change of time scale: measured in units of the time step, the gain of one
step varies by less than 13\% over all orders in $(0,1)$. This explains why a learned spectral density is
needed to turn the order into an effective nonlinearity. We propose a learning scheme in which the linear
readout is eliminated in closed form and is itself a smooth functional on the frequency axis, so that
what is trained is the continuous operator rather than its discretisation. In numerical experiments on
products of delayed inputs, a network of 20 units outperforms an untrained reservoir of comparable size,
keeps working when the orders move by only $\pm0.02$, and remains about one order of magnitude less
accurate than a trained gated recurrent network with a comparable number of parameters. The results show that
a state-dependent law of memory, acting on otherwise linear dynamics, is sufficient for nonlinear
computation, and they relate recent fractional and selective state-space architectures to the general
fractional calculus.
\end{abstract}

\section{Introduction}

A recurrent network computes by combining two ingredients: a memory, which keeps traces of past inputs,
and a nonlinearity, which combines those traces. In classical networks the two are entangled in the
same equation, $h_{t+1}=\tanh(Wh_t+Uu_t)$. Recent state-space models have pulled them partly apart. In
Mamba and its relatives the state evolves linearly between updates, and the nonlinearity has been moved
into input-dependent gates that set how fast each component of the state forgets and how inputs are
written and read \cite{gu2023mamba}. Feedback variants make the gates depend on the state itself
\cite{zattra2025coffee}. A step further in the same direction, FRAC replaces the exponential forgetting
of these models with power-law memory obtained from fractional dynamics and lets the fractional order
depend on the token \cite{kobyzev2026frac}. Neural differential equations of variable fractional order
\cite{cui2025nvofde}, attention-shaped fractional kernels \cite{kang2025fade}, distributed-order spiking
units \cite{ge2026dfsnn} and Hopfield networks of fractional-order neurons \cite{fazzino2021hopfield}
explore the same territory from the side of fractional calculus. In all these architectures, however,
the memory is only one of the places where the nonlinearity acts: either the vector field is itself
nonlinear, or the gates also rescale the input, the output and the time step.

This paper asks what happens if the separation is pushed to its limit. Suppose that every equation of the
network is linear, that the coupling between units is linear and fixed, and that the \emph{only}
nonlinear element is the order of differentiation of each equation, chosen at each instant by a smooth
function of the input and of the state. The dynamics are then still nonlinear, because the order
multiplies the memory of the system, but the nonlinearity has a very restricted form: one scalar per
equation that selects a law of memory from a fixed family. Can such a network learn nonlinear
input--output maps? And if it can, what does its structure tell us about the role that memory plays in
computation?

The question has a natural home in the theory of general fractional derivatives developed by Kochubei
\cite{kochubei2011} and Luchko \cite{luchko2021}. There, a fractional derivative is not tied to the power
law $t^{-\alpha}$: it is an operator of the form ``first derivative composed with a convolution'', and
the convolution kernel is required to have a partner kernel with which it convolves to one (the Sonine
condition). The class contains the Caputo derivative, tempered and distributed-order derivatives, and it
is a natural class of linear laws of memory for which both fundamental theorems of fractional calculus
hold. A network of linear general fractional systems whose kernels are selected by the state is
therefore a well-defined mathematical object, and the questions above can be asked with precision.

To make the object concrete we use the diffusive representation \cite{montseny1998,hinze2019}, which
rewrites a fractional operator as a continuum of first-order relaxations, one for each frequency
$\omega>0$. In this picture the memory of an equation is a bank of modes, the kernel is a density over the
bank, and the order is a single number that tilts this density towards slow or fast modes. In earlier work
we used the same representation as a phase space for fractional dynamics, to define and compute Lyapunov
exponents \cite{fazzino2026lyap,fazzino2026tutorial}, a problem we had first approached with a
semi-analytical method \cite{caponetto2013}, and to study the fractional Hopf bifurcation
\cite{fazzino2026hopf}; here it becomes the architecture of a learning machine.

The contributions of the paper are the following.
\begin{itemize}
\item We define the network precisely: linear units with general fractional derivatives, a Sonine kernel
for every equation, and an order that depends on input and state. Because the order varies in time, the
definition requires a choice of how past inputs are weighted when the order changes; we adopt the
``injection'' form, in which every input is stored with the order that holds when it arrives.
\item We prove three structural properties: equilibria independent of the orders, a linearisation in
which the orders appear frozen, and a step gain that, for a pure fractional kernel and time measured in
units of the step, is almost insensitive to the order. The third property is elementary, but it explains
why a learned spectral density is needed to turn the order into a working nonlinearity.
\item We give a learning scheme matched to this structure: the readout is removed in closed form, as in
variable projection \cite{golub1973}, and is itself a smooth functional over the frequency axis, so that
the trained object is the continuous operator rather than a particular set of modes.
\item We show numerically that the network learns products of delayed inputs, that it does better than an
untrained reservoir of the same size, that it still works when the orders move only by $\pm0.02$, and
that it is less accurate than a trained gated recurrent unit with a comparable number of parameters.
\end{itemize}

Two implications seem to us worth stating at the outset. The first concerns interpretation: for any frozen
value of its order every unit is an explicit linear operator with a known kernel, and the whole nonlinear
behaviour of a trained network is contained in time-varying exponents that can be read and plotted. The
second concerns the place of memory in computation: since equilibria are linear and independent of the
orders, a network of this kind cannot store information in attractors; it computes only while the input
is changing, by deciding how long each piece of information should be remembered. This is also the regime
in which selective state-space models operate, and our construction isolates, within the general
fractional calculus, the part of their mechanism that consists in modulating a law of memory.

The paper is organised as follows. Section~\ref{sec:background} recalls the general fractional calculus
and the diffusive representation. Section~\ref{sec:network} defines the network. Section~\ref{sec:props}
proves the structural properties. Section~\ref{sec:learning} describes the learning scheme,
Section~\ref{sec:experiments} the experiments, and Section~\ref{sec:discussion} discusses implications
and limits.

\section{Background: general fractional derivatives and the diffusive representation}
\label{sec:background}

\paragraph{From Caputo to Kochubei--Luchko.}
For $0<\alpha<1$ the Caputo derivative of a function $x(t)$ is
\begin{equation}
{}^{C}\!D^{\alpha}x(t)=\frac{1}{\Gamma(1-\alpha)}\int_0^t (t-s)^{-\alpha}\,\dot x(s)\dd s ,
\label{eq:caputo}
\end{equation}
that is, the velocity $\dot x$ averaged over the past with a weight that decays as a power of the
elapsed time. The equation ${}^{C}\!D^{\alpha}x=v$ is equivalent to the integral equation
\begin{equation}
x(t)=x(0)+\int_0^t K(t-s)\,v(s)\dd s,\qquad K(t)=\frac{t^{\alpha-1}}{\Gamma(\alpha)} ,
\label{eq:integral}
\end{equation}
which says that $x$ accumulates its input $v$ with a memory kernel $K$. The two kernels
$k(t)=t^{-\alpha}/\Gamma(1-\alpha)$ in \eqref{eq:caputo} and $K(t)=t^{\alpha-1}/\Gamma(\alpha)$ in
\eqref{eq:integral} satisfy $\int_0^t k(t-s)K(s)\dd s=1$ for all $t>0$, and it is this identity that
makes the derivative and the integral inverse to each other.

Kochubei \cite{kochubei2011} and Luchko \cite{luchko2021} turned this observation into a definition.
A pair of locally integrable kernels $(k,K)$ is a \emph{Sonine pair} if
\begin{equation}
(k*K)(t)=\int_0^t k(t-s)\,K(s)\dd s=1\qquad\text{for all }t>0 .
\label{eq:sonine}
\end{equation}
The general fractional derivative of Caputo type with kernel $k$ is
${}^{C}\!D_{(k)}x(t)=\int_0^t k(t-s)\,\dot x(s)\dd s$, and the equation ${}^{C}\!D_{(k)}x=v$ is then
equivalent to \eqref{eq:integral} with $K$ the partner of $k$. In this paper we work with the integral
form \eqref{eq:integral} and specify the memory kernel $K$; the corresponding derivative is guaranteed
to exist when $K$ belongs to a Sonine pair. A convenient sufficient condition is that $K$ be
\emph{completely monotone} (positive, decreasing, convex, and so on for all derivatives) and unbounded
at $t=0$: such a kernel has a completely monotone Sonine partner (see Chapter~5 of \cite{gripenberg1990}),
so that the corresponding general fractional derivative exists and has a kernel of the same type.

\paragraph{The diffusive representation.}
By Bernstein's theorem, a completely monotone kernel is a superposition of decaying exponentials,
\begin{equation}
K(t)=\int_0^\infty \rho(\omega)\,e^{-\omega t}\dd\omega ,\qquad \rho\ge 0 .
\label{eq:bernstein}
\end{equation}
For the Caputo kernel the density is $\rho_\alpha(\omega)=\frac{\sin(\pi\alpha)}{\pi}\,\omega^{-\alpha}$.
Inserting \eqref{eq:bernstein} into \eqref{eq:integral} and exchanging the integrals gives
\begin{equation}
x(t)=x(0)+\int_0^\infty \zeta(\omega,t)\dd\omega,\qquad
\partial_t\zeta(\omega,t)=-\omega\,\zeta(\omega,t)+\rho(\omega)\,v(t),\qquad \zeta(\omega,0)=0 .
\label{eq:diffusive}
\end{equation}
Each $\zeta(\omega,\cdot)$ is a first-order relaxation with rate $\omega$, fed by the input $v$ with
weight $\rho(\omega)$. Slow modes (small $\omega$) keep the input for a long time, fast modes forget it
almost at once. The fractional equation has become an infinite family of ordinary equations, and the
order $\alpha$ appears only in the weights $\rho_\alpha(\omega)$: a small $\alpha$ puts more weight on
slow modes and therefore produces a longer memory. This is the diffusive, or infinite-state,
representation of Montseny \cite{montseny1998}; it is exact, and in numerical work the integral over
$\omega$ is replaced by a quadrature on a logarithmic grid \cite{hinze2019}.

Equation \eqref{eq:diffusive} also shows what the general kernels of Kochubei and Luchko look like in
this language. Any positive density $\rho(\omega)$ defines a completely monotone kernel; it is a
Sonine kernel of the type considered here when $\rho(\omega)$ behaves as $\omega^{-\alpha}$ at large
frequencies, because this is what makes $K(t)\sim t^{\alpha-1}/\Gamma(\alpha)$ unbounded but
integrable near $t=0$. Multiplying the Caputo density by a positive profile $w(\omega)$ that tends to
one at the ends of the frequency axis therefore keeps us inside the class while allowing a different
distribution of memory over time scales. Other profiles give other familiar operators: a profile that
vanishes below a cut-off frequency gives a tempered power law, a sum of power laws gives a
multi-order or distributed-order operator. In this paper we use profiles that tend to one at both ends,
because the low-frequency behaviour is what guarantees the long memory used in
Proposition~\ref{prop:eq} below.

\section{The network}
\label{sec:network}

\paragraph{Units and coupling.}
The network has $M$ units, each made of $n$ scalar equations. We index equations by $(i,j)$, with $i$
the unit and $j=1,\dots,n$ the equation inside the unit, and collect the states $x_{ij}$ into a vector
$x\in\mathbb R^{Mn}$. Every equation has the integral form \eqref{eq:integral} with its own kernel,
\begin{equation}
x_{ij}(t)=x_{ij}(0)+\int_0^t K_{ij}(t,s)\,v_{ij}(s)\dd s,\qquad v=Ax+Bu ,
\label{eq:network}
\end{equation}
where $u(t)$ is the external input, $B$ a fixed input vector, and $A$ a fixed matrix. The diagonal
blocks $A_{ii}$ describe the internal dynamics of unit $i$; the off-diagonal blocks couple different
units. For given kernels, \eqref{eq:network} is linear in $x$ and $u$; the only nonlinearity enters
through the kernels, which depend on the state as we now explain.

\paragraph{State-dependent order.}
Each equation has an order $\alpha_{ij}(t)$ that is recomputed at every instant by a smooth gate,
\begin{equation}
\alpha_{ij}(t)=\alpha_{\min}+(\alpha_{\max}-\alpha_{\min})\,
\sigma\!\big(a_{ij}\,u(t)+b_{ij}\,\bar x_i(t)+c_{ij}\big),
\label{eq:gate}
\end{equation}
where $\sigma(z)=1/(1+e^{-z})$ is the logistic function and $\bar x_i=\frac1n\sum_j x_{ij}$ is the
output of unit $i$. The interval $[\alpha_{\min},\alpha_{\max}]\subset(0,1)$ fixes how far the order may
move. When $b_{ij}=0$ the order depends only on the input; when $b_{ij}\neq0$ it depends also on the
state, and the network becomes genuinely recurrent through its memory. The scalars
$a_{ij},b_{ij},c_{ij}$ are learned.

\paragraph{Spectral density.}
The kernel of equation $(i,j)$ is defined through its density on the frequency axis,
\begin{equation}
\rho_{ij}(\omega;\alpha)=w_{ij}(\omega)\,\frac{\sin(\pi\alpha)}{\pi}\,\omega^{-\alpha},\qquad
w_{ij}(\omega)=\exp\Big(\sum_{p=1}^{P}\theta_{ij,p}\,\varphi_p(\ln\omega)\Big).
\label{eq:density}
\end{equation}
The first factor is a learned profile, the second the Caputo density of order $\alpha$. The basis
functions $\varphi_p$ are $P$ Gaussian bumps in $\ln\omega$, centred inside a finite band of
frequencies, so $w_{ij}(\omega)\to1$ for $\omega\to0$ and $\omega\to\infty$. The profile is positive,
smooth, and does not alter the power-law behaviour at the two ends of the axis; for every fixed $\alpha$
the resulting kernel is completely monotone, unbounded at the origin, and therefore a Sonine kernel.
The coefficients $\theta_{ij,p}$ are learned. With $\theta=0$ the equation is a Caputo equation of
order $\alpha$.

\paragraph{How the order acts when it changes: the injection form.}
For constant order, \eqref{eq:network}--\eqref{eq:density} is a linear general fractional equation. When
the order varies in time, one has to decide what a variable-order memory means, and different choices
lead to different operators, as noted already by Lorenzo and Hartley \cite{lorenzo2002}. We use the
diffusive representation to make the choice explicit:
\begin{equation}
\partial_t\zeta_{ij}(\omega,t)=-\omega\,\zeta_{ij}(\omega,t)+\rho_{ij}\big(\omega;\alpha_{ij}(t)\big)\,v_{ij}(t),
\qquad x_{ij}(t)=x_{ij}(0)+\int_0^\infty\zeta_{ij}(\omega,t)\dd\omega .
\label{eq:injection}
\end{equation}
The order acts only on the weights with which the current input is written into the modes; once
written, the content of each mode relaxes at its own fixed rate. In words, every input is remembered
with the law of memory that was in force when it arrived. The kernel in \eqref{eq:network} is then
$K_{ij}(t,s)=\int_0^\infty\rho_{ij}(\omega;\alpha_{ij}(s))\,e^{-\omega(t-s)}\dd\omega$, which depends
on the time of writing $s$ through the order and on the elapsed time $t-s$ through the modes. The
alternative, in which the order reweights the modes at the time of reading, would let the present order
rewrite the whole past; we do not consider it here.

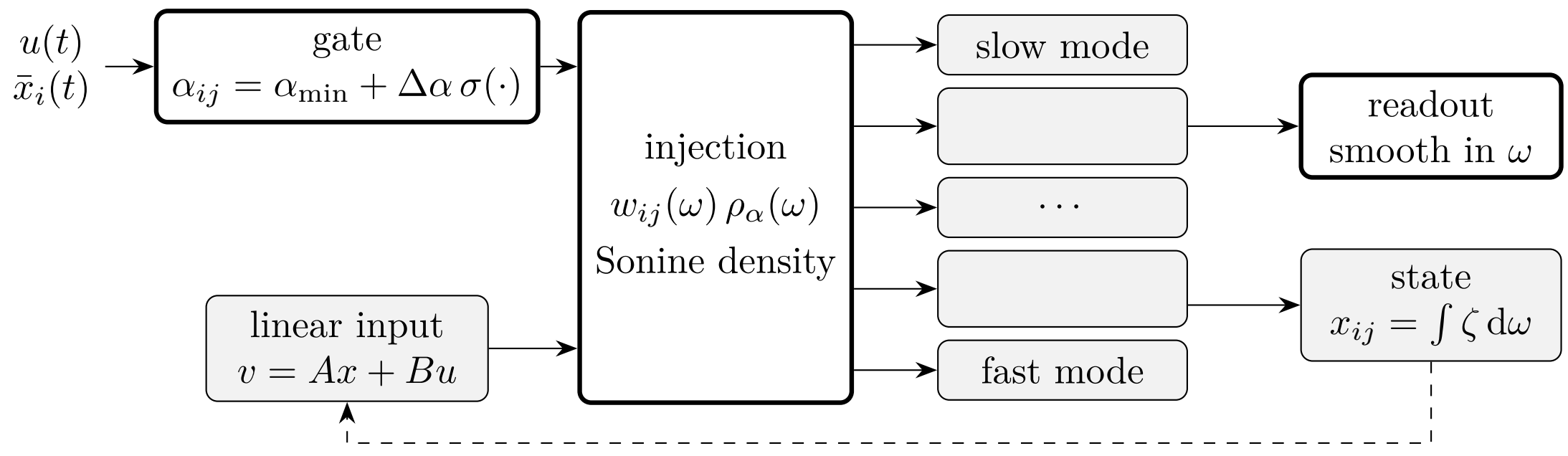
\begin{figure}[t]
\centering
\begin{tikzpicture}[font=\small,
  box/.style={draw,rounded corners=3pt,align=center,minimum height=0.9cm,inner sep=4pt},
  fixed/.style={box,fill=gray!10},
  learn/.style={box,fill=black!0,very thick},
  >={Stealth[length=2mm]}]
\node[learn,minimum width=2.6cm] (gate) at (0,2.2) {gate\\$\alpha_{ij}=\alpha_{\min}+\Delta\alpha\,\sigma(\cdot)$};
\node[fixed,minimum width=2.6cm] (v) at (0,-0.4) {linear input\\$v=Ax+Bu$};
\node[left=0.45cm of gate,align=center] (z) {$u(t)$\\$\bar x_i(t)$};
\draw[->] (z) -- (gate);
\node[learn,minimum width=2.5cm,minimum height=3.6cm] (inj) at (3.4,0.9) {injection\\[3pt]$w_{ij}(\omega)\,\rho_{\alpha}(\omega)$\\[3pt]Sonine density};
\draw[->] (gate.east) -- (gate.east -| inj.west);
\draw[->] (v.east) -- (v.east -| inj.west);
\foreach \y/\lab/\nm in {2.4/{slow mode}/a,1.65/{\strut}/b,0.9/{$\cdots$}/c,0.15/{\strut}/d,-0.6/{fast mode}/e}{
  \node[fixed,minimum width=2.3cm,minimum height=0.55cm] (m\nm) at (6.6,\y) {\lab};
  \draw[->] (inj.east |- m\nm) -- (m\nm.west);
}
\node[learn,minimum width=2.4cm] (read) at (10.0,1.65) {readout\\smooth in $\omega$};
\node[fixed,minimum width=2.4cm] (state) at (10.0,0.0) {state\\$x_{ij}=\int\zeta\dd\omega$};
\draw[->] (7.75,1.65) -- (read.west);
\draw[->] (7.75,0.0) -- (state.west);
\draw[->,dashed] (state.south) -- ++(0,-0.75) -| (v.south);
\node[font=\footnotesize] at (5,-1.75) {fed back through $A$ and to the gate};
\end{tikzpicture}
\caption{One equation of the network. The gate computes the order ($\Delta\alpha=\alpha_{\max}-\alpha_{\min}$) from the input and from the output of
the unit. The linear input $v$ is injected into a bank of relaxing modes with weights given by the
learned profile $w_{ij}$ times the fractional density of the current order. The modes are summed into
the state, which closes the linear loop through $A$, and are read individually by a readout that is a
smooth function of the frequency. Thick frames mark learned components.}
\label{fig:architecture}
\end{figure}

Figure~\ref{fig:architecture} summarises the architecture. Fixed parts are the coupling $A$, the input
vector $B$ and the frequencies of the modes; learned parts are the gate coefficients, the spectral
profiles, and the readout described in Section~\ref{sec:learning}.

\paragraph{Discretisation.}
For computation the frequency axis is replaced by $K$ logarithmically spaced nodes
$\omega_1<\dots<\omega_K$ with trapezoidal weights $q_k$, plus one lumped mode that collects all
frequencies below $\omega_1$ and behaves as a pure integrator. With a time step $h$ and the input held
constant over a step, each mode is integrated exactly:
\begin{equation}
\begin{aligned}
\zeta_{ij,k}(t+h)&=e^{-\omega_k h}\,\zeta_{ij,k}(t)+\frac{1-e^{-\omega_k h}}{\omega_k}\,q_k\,
\rho_{ij}\big(\omega_k;\alpha_{ij}(t)\big)\,v_{ij}(t),\\
x_{ij}(t)&=x_{ij}(0)+\textstyle\sum_k\zeta_{ij,k}(t).
\end{aligned}
\label{eq:discrete}
\end{equation}
The decay of each mode is integrated exactly, so the fast modes impose no restriction on $h$; the coupling
through $A$ is treated explicitly, which requires $h$ to be small compared with the time scales of $A$.
The only nonlinear operation in \eqref{eq:discrete} is the evaluation of the weights at the current
order.

\section{Structural properties}
\label{sec:props}

The three results of this section use only the linearity of \eqref{eq:network}, the positivity of the
densities and the boundedness of the orders. The two propositions hold for the continuous model and for
its discretisation, in which the lumped integrator plays the role of the low-frequency modes.

\begin{proposition}[equilibria do not depend on the orders]\label{prop:eq}
Let the input be constant, $u(t)\equiv u_*$, and let the orders stay in
$[\alpha_{\min},\alpha_{\max}]\subset(0,1)$. If a solution of \eqref{eq:injection} converges to a constant
state $x_*$, then $Ax_*+Bu_*=0$. In particular, if $A$ is invertible the only possible equilibrium is
$x_*=-A^{-1}Bu_*$, whatever the gates and the spectral profiles.
\end{proposition}

\begin{proof}
If $x(t)\to x_*$ then $v(t)\to v_*=Ax_*+Bu_*$. Suppose $v_{*,ij}\neq0$ for some equation and take $T$ such
that $v_{ij}(s)$ has the sign of $v_{*,ij}$ and modulus at least $|v_{*,ij}|/2$ for $s\ge T$. Since the
profile $w_{ij}$ tends to one at low frequency, for $\omega$ below some $\omega_0$ the density satisfies
$\rho_{ij}(\omega;\alpha)\ge c\,\omega^{-\alpha_{\min}}$ with $c>0$ for every admissible $\alpha$. The
modes below $\omega_0$ then contribute to $x_{ij}(t)-x_{ij}(T)$ a quantity of fixed sign and modulus at
least
\[
\frac{|v_{*,ij}|}{2}\,c\int_0^{\omega_0}\omega^{-\alpha_{\min}}\,\frac{1-e^{-\omega(t-T)}}{\omega}\dd\omega ,
\]
which grows without bound as $t\to\infty$ because $\omega^{-\alpha_{\min}-1}$ is not integrable at
the origin. The remaining modes stay bounded. Hence $x_{ij}(t)$ diverges, against the assumption.
\end{proof}

The meaning of Proposition~\ref{prop:eq} is that, at rest, the network is a linear map: the gates and
the spectral densities cannot shape the steady state. When $A$ is invertible the network has a single
equilibrium for each constant input, no multistability, and no memory stored in attractors. Whatever nonlinear computation it
performs takes place while the input changes.

\begin{proposition}[the linearisation sees frozen orders]\label{prop:lin}
Let $x_*$ be an equilibrium for the constant input $u_*$, and let $\alpha_*$ be the orders produced by
the gates at $(u_*,x_*)$. The linearisation of \eqref{eq:injection} at $x_*$ is the linear general
fractional system with constant orders $\alpha_*$,
\[
\partial_t\delta\zeta_{ij}=-\omega\,\delta\zeta_{ij}+\rho_{ij}(\omega;\alpha_{*,ij})\,(A\,\delta x)_{ij}.
\]
\end{proposition}

\begin{proof}
The injected term is $\rho_{ij}(\omega;\alpha_{ij})\,v_{ij}$. Its first-order variation is
\[
\rho_{ij}(\omega;\alpha_{*,ij})\,\delta v_{ij}+\partial_\alpha\rho_{ij}(\omega;\alpha_{*,ij})\,v_{*,ij}\,\delta\alpha_{ij}.
\]
By Proposition~\ref{prop:eq}, $v_*=0$, so the second term vanishes and $\delta v=A\,\delta x$.
\end{proof}

The dependence of the orders on the state is therefore invisible to a first-order analysis: it changes
neither the position of the equilibria nor the linearised dynamics around them, and it matters only along
trajectories that move away from equilibrium, which is precisely where a network processing a
time-varying input lives. Turning the linearisation into a statement about local asymptotic stability
requires a linearised stability theorem; such a theorem is available for Caputo systems of a
single order \cite{cong2016}, while for general kernels and orders that vary along the trajectory this remains to be
established.

The third property explains why the spectral profile matters. It concerns the units in which time is
measured. The Caputo kernel $t^{\alpha-1}/\Gamma(\alpha)$ in \eqref{eq:integral} is written for a
dimensionless time; with a physical time unit $\tau$ it reads $\tau^{-\alpha}t^{\alpha-1}/\Gamma(\alpha)$,
and in a discrete-time network the natural unit is the step, $\tau=h$. Consider a single equation with
the pure Caputo density ($w\equiv1$), at rest, receiving a constant input $v$ for one step. From
\eqref{eq:integral},
\begin{equation}
x(h)-x(0)=v\int_0^h\frac{(h-s)^{\alpha-1}}{\tau^{\alpha}\,\Gamma(\alpha)}\dd s
=\Big(\frac{h}{\tau}\Big)^{\alpha}\frac{v}{\Gamma(1+\alpha)} .
\label{eq:stepgain}
\end{equation}
The order enters the step gain in two ways: through the factor $(h/\tau)^\alpha$, which is the effect of
a change of time scale, and through $1/\Gamma(1+\alpha)$, which is all that remains when time is measured
in units of the step.

\begin{lemma}[with $\tau=h$ the order acts on timing, not on amplitude]\label{lem:gain}
For $\tau=h$ the step gain $1/\Gamma(1+\alpha)$ lies between $1$ and $1/\Gamma(1.4616)\approx1.129$ for
every $\alpha\in[0,1]$. Over the same range the weight $\omega^{-\alpha}$ with which the input enters a
mode of frequency $\omega$ (in units of $1/h$) changes by the factor $\omega^{-1}$, that is by orders of
magnitude for modes much slower or much faster than the step.
\end{lemma}

\begin{proof}
$\Gamma$ is log-convex on $(0,\infty)$ with minimum $\approx0.8856$ at $\approx1.4616$, and
$\Gamma(1)=\Gamma(2)=1$. The second statement is immediate.
\end{proof}

\begin{figure}[t]
\centering
\includegraphics[width=\linewidth]{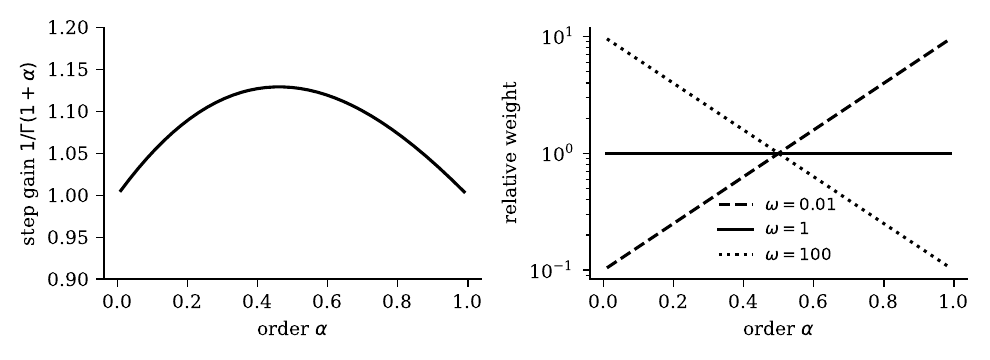}
\caption{Left: gain of one time step, $1/\Gamma(1+\alpha)$, for a pure Caputo equation with time
measured in units of the step. Right: weight
$\omega^{-\alpha}$ of a slow ($\omega=0.01$), intermediate ($\omega=1$) and fast ($\omega=100$) mode,
normalised at $\alpha=0.5$. The order redistributes the input among time scales almost without
changing its total.}
\label{fig:gain}
\end{figure}

Figure~\ref{fig:gain} shows the two curves. When the time unit is the step, changing the order hardly
changes how much of a new input enters the state; it changes how this amount is distributed among the
modes, and therefore at which time scales it will later be felt. A change of order thus behaves much like
a change of clock, and in fractional systems a change of clock is not a neutral operation: it alters
orbits and bifurcations \cite{fazzino2026clock}. For the network, the consequence is practical. If the
readout looks only at the state $x_{ij}$, which is the sum of the modes, the effect of the order is weak at
short times. Two ingredients make it visible: a profile $w_{ij}(\omega)$ that weights the modes
unequally, so that redistributing an input among them changes the total (concentrating the weight near a
frequency $\omega_0$ has the same effect as choosing the time unit $\tau=1/\omega_0$, but frequency by
frequency), and a readout that sees the modes separately. Both are part of the network defined above.

\section{Learning}
\label{sec:learning}

\paragraph{Readout as a smooth functional of the diffusive variable.}
The output of the network is a linear functional of the modes,
\begin{equation}
y(t)=\sum_{i,j}\int_0^\infty d_{ij}(\omega)\,\zeta_{ij}(\omega,t)\dd\omega+d_0,
\qquad d_{ij}(\omega)=\sum_{p=1}^{P}D_{ij,p}\,\varphi_p(\ln\omega),
\label{eq:readout}
\end{equation}
where the readout profiles $d_{ij}$ are expanded on the same smooth basis used for the spectral
densities. In discrete form the network therefore exposes $P$ features per equation,
\[
F_{ij,p}(t)=\textstyle\sum_k\varphi_p(\ln\omega_k)\,\zeta_{ij,k}(t),
\]
and the output is $y=F D$. The choice
has a practical and a conceptual reason. Practically, it reduces the number of features from the number
of modes to $P$ per equation. Conceptually, \eqref{eq:readout} is defined on the continuous frequency
axis: refining the grid of modes changes the quadrature, not the model. What is learned is the operator,
not its discretisation.

\paragraph{Variable projection.}
Let $\Theta$ collect the gate coefficients $(a,b,c)$ and the spectral coefficients $\theta$. For fixed
$\Theta$ the features $F(\Theta)$ are determined by simulating the network, and the best readout is a
ridge regression,
\begin{equation}
D^*(\Theta)=\arg\min_D\ \frac1T\|F(\Theta)D-y_{\mathrm{target}}\|^2+\lambda\|D\|^2
=\big(F^\top F/T+\lambda I\big)^{-1}F^\top y_{\mathrm{target}}/T .
\label{eq:ridge}
\end{equation}
Substituting $D^*(\Theta)$ eliminates the readout and leaves a function of $\Theta$ alone,
\[
J(\Theta)=\tfrac1T\|F(\Theta)D^*(\Theta)-y_{\mathrm{target}}\|^2 .
\]
This is the idea of variable projection \cite{golub1973}: parameters that enter the model linearly are
eliminated analytically, and the optimiser works only on the nonlinear ones. Because $D^*(\Theta)$ is an
explicit, differentiable function of the features, the gradient of $J$ is computed exactly, including the
dependence that passes through $D^*$. The readout is never learned by gradient descent, and the
optimiser deals only with the few parameters that shape the memory: eight per equation in our networks
(three for the gate, five for the profile). In practice the features are standardised (centred and
divided by their standard deviation over the training segment) before \eqref{eq:ridge} is applied, which
makes a single value of $\lambda$ adequate for all features.

\paragraph{Gradient with respect to the order.}
All the dependence of the features on the gate parameters passes through the injection weights. The
derivative of the Caputo density with respect to the order has the closed form
\begin{equation}
\frac{\partial\rho(\omega;\alpha)}{\partial\alpha}=\rho(\omega;\alpha)\,\big(\pi\cot(\pi\alpha)-\ln\omega\big).
\label{eq:drho}
\end{equation}
The factor $-\ln\omega$ is negative for modes faster than the step ($\omega>1$ in units of $1/h$) and
positive for slower ones: increasing the order moves
weight from slow to fast modes. The gradient of $J$ with respect to a gate coefficient is a time
integral of the adjoint (back-propagated) error, projected on the modes with the weights \eqref{eq:drho}.
It tells each equation, at each instant, whether the input just received should have been written into
longer or shorter memory. In practice we compute it by reverse-mode automatic differentiation through
the exact update \eqref{eq:discrete}; the adjoint of a diffusive system is again a diffusive system,
run backwards in time, whose modes decay with the same rates as in the forward pass.

\paragraph{Procedure.}
Training uses a single long input sequence. The first part of the sequence is discarded as transient,
the last 20\% of the remainder is set aside for selecting the iterate, and the rest is used to compute
$D^*$ and $J$. The parameters $\Theta$ are updated with the Adam optimiser; at every iteration the
network is simulated, the readout is recomputed by \eqref{eq:ridge}, and the gradient of $J$ is taken.
The iterate with the smallest error on the held-out segment is kept; its readout is finally refitted on
the whole training sequence with a ridge parameter chosen on the same segment.

\section{Numerical experiments}
\label{sec:experiments}

\paragraph{Task.}
The input $u(t)$ is a sequence of independent values uniformly distributed in $[-1,1]$, one per time
step. The target is the product of two past inputs,
\begin{equation}
y_{\mathrm{target}}(t)=u(t-1)\,u(t-1-\ell),\qquad \ell\in\{1,2,4\}.
\label{eq:task}
\end{equation}
Because the inputs have zero mean and are independent, the target is uncorrelated with every linear
function of past inputs: the best linear predictor, with any memory, has a normalised error of one.
Any improvement below one measures nonlinear computation. Larger $\ell$ require the product of inputs
farther apart in time. The error is the mean squared error on an independent test sequence of 1200
steps, divided by the variance of the target.

\paragraph{Network.}
We use $M=20$ units with $n=3$ equations each, so 60 equations and 60 orders. Every unit is different:
the internal block is $A_{ii}=-\mathrm{diag}(d)+(S-S^\top)$, with $d$ uniform in $[0.2,1.5]$ and $S$ a
Gaussian matrix with standard deviation $0.8$, so that each unit has its own relaxation rates and oscillations; units are coupled by a
sparse random matrix (10\% density) scaled so that the symmetric part of $A$ remains negative definite,
and $A$ is finally multiplied by $0.1$ to give internal time scales of tens of steps. The modes are
$K=12$ frequencies from $10^{-3}$ to $10^{2}$ plus the lumped low-frequency mode; the time step is
$h=1$. The spectral profiles use $P=5$ Gaussian bumps in $\ln\omega$ of width $1.2$, centred between
$10^{-3}$ and $10$. The gates take both the input and the output of the unit (Eq.~\ref{eq:gate}). In
total the network has 480 trained parameters plus a readout of 301 coefficients computed in closed form.
Training sequences have 1500 steps; Adam runs for 150 iterations with learning rate $0.02$. Every
configuration is run on two independent realisations of network and data.

\paragraph{Baselines.}
A \emph{linear control} is the same network with all orders fixed at $0.6$; it is a linear general
fractional system and cannot compute products. A \emph{reservoir} is an echo state network
\cite{jaeger2001} of 60 hyperbolic-tangent units with spectral radius $0.9$, random input and bias
weights, untrained internal connections and a ridge readout. A \emph{gated recurrent unit}
\cite{cho2014} with 12 hidden units (553 parameters including its readout) is trained end-to-end on
the same data, with the same held-out segment, by Adam (600 iterations, learning rate $0.01$). The
reservoir and the gated unit answer two different questions. The reservoir has a fixed nonlinear
dynamics and only a trained readout, so it tells whether the nonlinearity produced by the orders is
strong enough to be useful. The gated unit is fully trained, like our network, so it is the fair
comparison for accuracy.

\paragraph{Results.}
Table~\ref{tab:main} collects the errors for orders free to move in $[0.3,0.95]$. The linear control
confirms that the nonlinearity comes entirely from the orders: with orders fixed, the error is one. With
state-dependent orders the network learns the products: at $\ell=1$ and $\ell=2$ the error is one to two
percent of the target variance, five to ten times smaller than that of the reservoir. At $\ell=4$ the reservoir
essentially fails, while our network still captures most of the product in one realisation and about
half in the other. Reading all thirteen modes of each equation individually (the twelve nodes and the
lumped mode) instead of the smooth profiles \eqref{eq:readout} gives nearly the same error (0.022 and
0.027 at $\ell=2$) with 780 features instead of 300, which supports the choice of a smooth readout. The trained gated recurrent unit is better than
our network by about one order of magnitude.

\begin{table}[t]
\centering
\caption{Normalised test error for the product $u(t-1)\,u(t-1-\ell)$; two independent realisations per
entry. Orders in $[0.3,0.95]$. A value of one means no nonlinear computation.}
\label{tab:main}
\begin{tabular}{lccc}
\toprule
 & $\ell=1$ & $\ell=2$ & $\ell=4$\\
\midrule
linear control (orders fixed at 0.6) & -- & 1.02 / 1.02 & -- \\
reservoir, 60 tanh units & 0.066 / 0.050 & 0.17 / 0.19 & 0.73 / 0.92\\
present network & 0.012 / 0.008 & 0.018 / 0.020 & 0.44 / 0.12\\
gated recurrent unit, 553 parameters & -- & 0.002 / 0.003 & 0.008 / 0.005\\
\bottomrule
\end{tabular}
\end{table}

\begin{figure}[t]
\centering
\includegraphics[width=\linewidth]{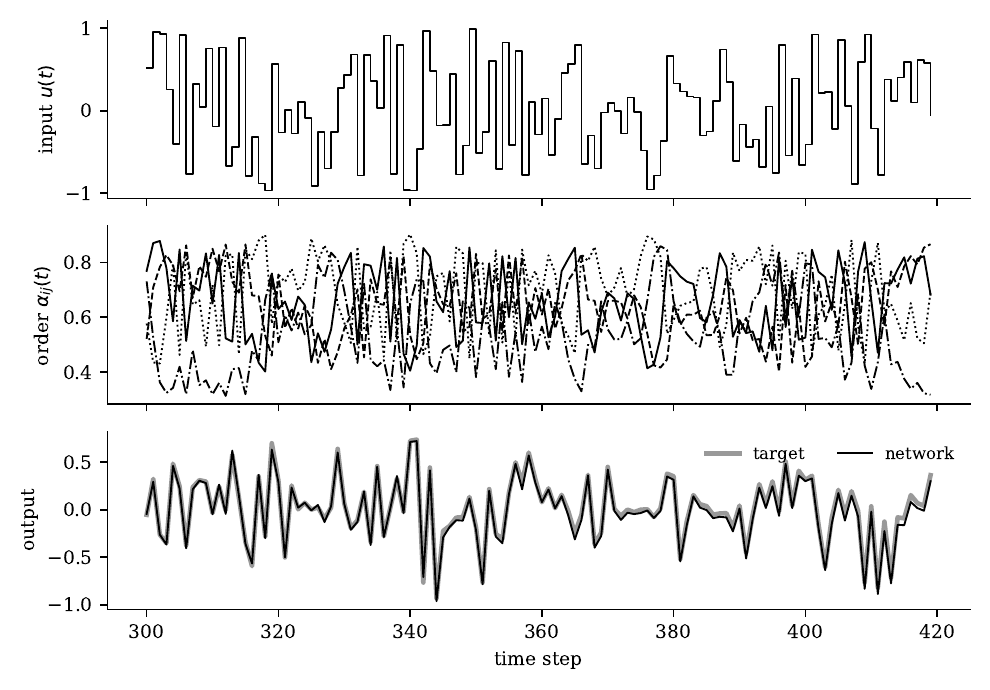}
\caption{A trained network on the test sequence ($\ell=2$). Top: input. Middle: orders of the four
equations whose order varies most. Bottom: target $u(t-1)u(t-3)$ (thick grey) and network output
(black).}
\label{fig:dynamics}
\end{figure}

\begin{figure}[t]
\centering
\includegraphics[width=\linewidth]{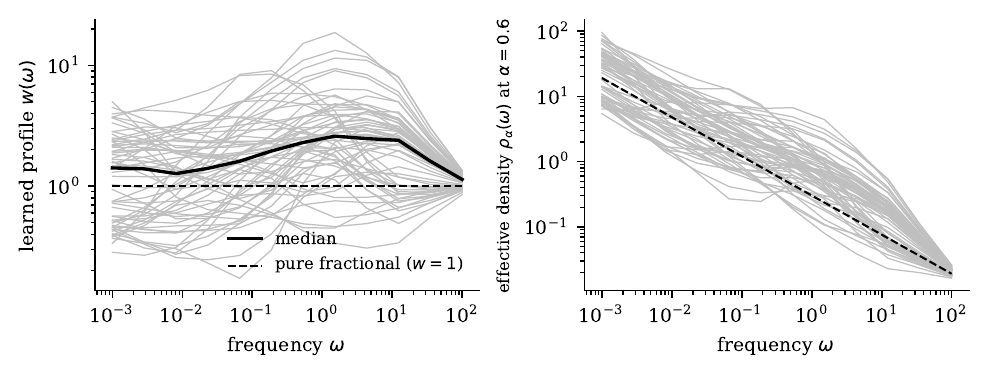}
\caption{Left: learned profiles $w_{ij}(\omega)$ of the 60 equations at the nodes of the bank (grey),
their median (black) and the pure fractional case $w=1$ (dashed). Right: resulting densities
$\rho_{ij}(\omega;0.6)$ compared with the Caputo density of the same order (dashed).}
\label{fig:spectra}
\end{figure}

Figure~\ref{fig:dynamics} shows a trained network at work on the test sequence for $\ell=2$. The orders
of the most active equations move over most of the allowed interval, step by step, in response to the
input and to the state; the output follows the target closely. Figure~\ref{fig:spectra} shows the learned
spectral profiles. They differ from equation to equation, but on average they raise the weight of the
intermediate frequencies, between $10^{-1}$ and $10$, by a factor of about two and a half, and they return
towards one at the fast end, as imposed by the construction. The effective densities remain
close to power laws on the scale of the whole axis: the network does not abandon fractional memory, it
reshapes it locally around the time scales of the task.

\paragraph{Small excursions of the order.}
Table~\ref{tab:narrow} repeats the experiment with the orders confined to $[0.7,0.9]$ and to
$[0.78,0.82]$. The network keeps learning. With an excursion of $\pm0.1$ the error at $\ell=1$ drops to
$0.003$; with only $\pm0.02$ around $0.8$ the errors are $0.017$--$0.066$, still better than the
reservoir of Table~\ref{tab:main}. One realisation at $\ell=2$ with $[0.7,0.9]$ is markedly worse than
the other (0.156); with two realisations we cannot say whether this reflects that particular random
network or a systematic difficulty.

\begin{table}[t]
\centering
\caption{Normalised test error with narrow ranges of the order; two realisations per entry.}
\label{tab:narrow}
\begin{tabular}{lcc}
\toprule
range of the order & $\ell=1$ & $\ell=2$\\
\midrule
$[0.3,0.95]$ & 0.012 / 0.008 & 0.018 / 0.020\\
$[0.7,0.9]$ & 0.003 / 0.003 & 0.032 / 0.156\\
$[0.78,0.82]$ & 0.049 / 0.017 & 0.066 / 0.027\\
\bottomrule
\end{tabular}
\end{table}

\section{Discussion}
\label{sec:discussion}

\paragraph{What the experiments show.}
A network whose equations are all linear, whose coupling is fixed, and whose only nonlinear element is
the order of differentiation can learn nonlinear functions of its input history. It does so better than
an untrained nonlinear reservoir of the same size, and less well than a fully trained gated network. The
first comparison answers the question we asked: a law of memory chosen at each instant is a
nonlinearity strong enough to compute with. The second shows that, at this scale and on these tasks, it
is not a better nonlinearity than the usual ones. Our network trains 480 parameters on top of fixed random
dynamics, against 553 for the gated unit, so the difference in accuracy is not a matter of parameter
count.

\paragraph{Where the nonlinearity lives.}
Propositions~\ref{prop:eq} and~\ref{prop:lin} say that the steady states and the linearised dynamics
around them are those of a linear system. Lemma~\ref{lem:gain} adds that, for pure fractional kernels and
time measured in units of the step, the order hardly changes the immediate response to an input. The nonlinearity therefore lives in a specific place: in the way
each input is distributed among time scales, and in the later consequences of that distribution. This
has a practical consequence for design: the readout must see the time scales separately and the kernel
must weight them unequally, otherwise the effect of the order cancels in the sum. It also has a
conceptual consequence. In a network of this kind, computing means deciding, for each piece of
information, how long it should be remembered.

\paragraph{Relation to selective state-space models.}
The gates of Mamba \cite{gu2023mamba} and FRAC \cite{kobyzev2026frac} change, among other things, the
rate at which the state forgets; the feedback gates of \cite{zattra2025coffee} do the same with the state
as input. In the language of Section~\ref{sec:background}, part of what these gates do is to modulate
the kernel of a linear memory according to the input or the state. Our construction isolates this part,
removes every other nonlinearity, and places it inside the general fractional calculus. The reward is a model whose units,
for frozen orders, are explicit linear operators with Sonine kernels and whose nonlinear behaviour is entirely described by
time-varying exponents that can be read and plotted, as in Figure~\ref{fig:dynamics}. The neural
fractional models of \cite{cui2025nvofde,kang2025fade,ge2026dfsnn} also learn orders or kernels, but
within nonlinear vector fields; the question of what the memory alone can compute does not arise there.

\paragraph{Limits.}
Three limits should be stated clearly. First, the experiments are small: one family of tasks, networks of 60
equations, two realisations per configuration; we have not tested long sequences or real data. Second, the results with narrow ranges of the order rely on
exact arithmetic: with an excursion of $\pm0.02$ the differences produced by the order are small, and
the readout amplifies them. Noise of comparable size on the states would hide them. This does not
affect the mathematical model, which is our subject, but it matters for any physical realisation and for
any use with noisy inputs. Third, two theoretical questions remain open. We have not proved a
universality theorem. The network is a linear system whose input weights are smooth functions of input
and state, which makes it a generalised bilinear system; bilinear systems can approximate continuous
causal input--output maps on finite time intervals \cite{sussmann1976}, and operators with fading memory
are approximated by Volterra series \cite{boyd1985}. Whether the same holds when the nonlinearity is
restricted to the family $\omega^{-\alpha}$ needs a dedicated argument. And we have not established global stability for
time-varying orders that differ from equation to equation; the negative-definite symmetric part of $A$
used here is a natural candidate for a sufficient condition.

\section{Conclusions}

We have studied neural networks built from linear general fractional systems in which the only
nonlinear element is the order of differentiation, chosen at each instant from the input and the state.
The diffusive representation gives the model a precise form, the Sonine condition keeps every equation a
genuine general fractional equation, and the injection form fixes the meaning of a variable order.
The equilibria of such networks are linear and independent of the orders, their linearisation contains
the orders only through their frozen values, and, with time measured in units of the step, the order of
a pure fractional kernel acts on the timing of the response rather than on its amplitude. Learning works when the kernels carry a learned spectral profile and the readout
sees the time scales separately; with variable projection and a readout defined on the continuous
frequency axis, the network learns products of delayed inputs better than a reservoir of the same size,
even when the orders move by only $\pm0.02$, although a trained gated network of comparable size remains
more accurate. The main message is that a state-dependent law of memory, acting on otherwise linear
dynamics, can carry the nonlinearity of a learning machine. Whether this makes such networks preferable in some
application, for instance in the identification of physical systems whose memory is known to change
with their state, is the natural question for further study.

\section*{Code availability}
The Python code that reproduces every table and figure of this paper, with instructions, is distributed
with the article as ancillary material. All computations use double precision on a single CPU core and
take less than one hour in total.

\end{document}